%% file: paper3033.tex
\documentclass[runningheads]{llncs}
\usepackage{graphicx}
\usepackage{subfig}
\usepackage{gensymb}
\usepackage[utf8]{inputenc}
\usepackage{amsfonts}
\usepackage{amsmath}

\newcommand{\vett}[1]{\boldsymbol{\mathbf{#1}}}
\newcommand{\qot}[1]{``#1''}
\newcommand{\mm}{mm}
\newcommand{\ms}{ms}

\newcommand{\smallvspace}{\vspace{0.4\baselineskip}}

\newcommand{\dotp}[2]{\langle #1, #2 \rangle}

\begin{document}
\title{DISA: DIfferentiable Similarity Approximation for Universal Multimodal Registration}
\titlerunning{DISA: Universal Multimodal Registration}

\author{Matteo Ronchetti\inst{1,2} \and Wolfgang Wein\inst{1} \and Nassir Navab \inst{2} \and Oliver Zettinig \inst{1} \and Raphael Prevost\inst{1}  }

\authorrunning{M. Ronchetti et al.}

\institute{ImFusion GmbH, M\"unchen, Germany
\and Computer Aided Medical Procedures (CAMP) Technische Universit\"at M\"unchen, Germany}
\maketitle

\begin{abstract}
Multimodal image registration is a challenging but essential step for numerous image-guided procedures. 
Most registration algorithms rely on the computation of complex, frequently non-differentiable similarity metrics to deal with the appearance discrepancy of anatomical structures between imaging modalities.
Recent Machine Learning based approaches are limited to specific anatomy-modality combinations and do not generalize to new settings.
We propose a generic framework for creating expressive cross-modal descriptors that enable fast deformable global registration.  
We achieve this by approximating existing metrics with a dot-product in the feature space of a small convolutional neural network (CNN) which is inherently differentiable can be trained without registered data.
Our method is several orders of magnitude faster than local patch-based metrics and can be directly applied in clinical settings by replacing the similarity measure with the proposed one.
Experiments on three different datasets demonstrate that our approach generalizes well beyond the training data, yielding a broad capture range even on unseen anatomies and modality pairs, without the need for specialized retraining.
We make our training code and data publicly available.

\end{abstract}

\section{Introduction}
\input{introduction.tex}

\section{Approach}
\input{approach.tex}

\section{Method}
\input{training.tex}

\section{Experiments and Results}
\label{sec:experiments}
\input{experiments.tex}
\subsection{Deformable Registration of Abdominal US-CT and US-MR}
\label{sec:ctus}
\input{ctus.tex}

\section{Conclusion}
\input{conclusions.tex}

\newpage

\bibliographystyle{splncs04}
\bibliography{citations}

\end{document}

%% file: introduction.tex
Multimodal imaging has become increasingly popular in healthcare %
due to its ability to provide complementary anatomical and functional information.
However, to fully exploit its benefits, it is crucial to perform accurate and robust registration of images acquired from different modalities. Multimodal image registration is a challenging task due to differences in image appearance, acquisition protocols, and physical properties of the modalities. This holds in particular if ultrasound (US) is involved, and has not been satisfactorily solved so far.

While simple similarity measures directly based on the images' intensities such as sum of absolute (L1) or squared (L2) differences and normalized cross-correlation (NCC)~\cite{ncc} work well in monomodal settings, a more sophisticated approach is needed when intensities cannot be directly correlated.
Historically, a breakthrough in CT-MRI registration was achieved by Viola and Wells, who proposed Mutual Information~\cite{mutual_inf}.
Essentially, it abstracts the problem to the statistical concept of information theory and optimizes image-wide alignment statistics.
Broken down to patch level and inspired by ultrasound physics, the Linear Correlation of Linear Combination (LC$^2$) measure has shown to work well for US to MRI or CT registration~\cite{lc2,fuerst2014automatic}.
While dealing well with US specifics, it is not differentiable and expensive to compute.

As an alternative to directly assessing similarity on the original images, various groups have proposed to first compute intermediate representations, and then align these with conventional L1 or L2 metrics~\cite{entropy_images,mind}.
A prominent example is the Modality-Independent Neighbourhood Descriptor (MIND)~\cite{mind}, which is based on image self-similarity and has with minor adaptations (denoted MIND-SSC for self-similarity context) also been applied to US problems~\cite{ssc}.
Most recently, it has been shown that using 2D confidence maps-based weighting and adaptive normalization may further improve registration accuracy~\cite{self_similarity_vector}.
Yet, such feature descriptors are not expressive enough to cope with complex US artifacts and exhibit many local optima, therefore requiring closer initialization.

More recently, multimodal registration has been approached using various Machine Learning (ML) techniques. Some of these methods involve the utilization of Convolutional Neural Networks (CNN) to extract segmentation volumes from the source data, transforming the problem into the registration of label maps~\cite{muller2014,label_reg_2022}. Although these methods have demonstrated promising results, they are anatomy-specific and require the identification and labeling of structures that are visible in both modalities.
Other approaches are trained using ground truth registrations to directly predict the pose~\cite{orientation_estimation,montana2022} or to establish keypoint correspondences~\cite{keypoint_reg_2022,esteban_towards_2019}. However, these methods are not generalizable to different anatomies or modalities. Moreover, the paucity of precise and unambiguous ground truth registration, particularly in abdominal MR-US registration, exacerbates the overfitting problem, restricting generalization even within the same modality and anatomy.
It has furthermore been proposed in the past to utilize CNNs as a replacement for a similary metric. In \cite{haskins2019,sedghi2018}, the two images being registered are resampled into the same grid in each optimizer iteration, concatenated and fed into a network for similarity evaluation.
While such a measure can directly be integrated into existing registration methods, it still suffers from similar limitations in terms of runtime performance and modality dependance.

In contrast, we propose in this work to use a small CNN to approximate an expensive similarity metric with a straightforward dot product in its feature space.
Crucially, our method does not necessitate to evaluate the CNN at every optimizer iteration.
This approach combines ML and classical multimodal image registration techniques in a novel way, avoiding the common limitations of ML approaches: ground truth registration is not required, it is differentiable and computationally efficient, and generalizes well across anatomies and imaging modalities.

%% file: approach.tex
We formulate image registration as an optimization problem of a similarity metric~$s$ between the moving image  $M$ and the fixed image $F$ with respect to the parameters $\vett{\alpha}$ of a spatial transformation $T_\alpha : \Omega \rightarrow \Omega$.
Most multi-modal similarity metrics are defined as weighted sums of local similarities computed on patches.
Denoting $M \circ T_\alpha$ the deformed image, the optimization target can be expressed in the following way:
\begin{align}
\label{reg_obj}
f(\alpha) = \sum_{p \in \Omega} w(p) \ s(F[p], M \circ T_\alpha [p]) \, ,
\end{align}
where $w(p)$ is the weight assigned to the point $p$, $s(\cdot, \cdot)$ defines a local similarity and the $[\cdot]$ operator extracts a patch (or a pixel) at a given spatial location.
This definition encompasses SSD but also other more elaborate metrics like $LC^2$ or MIND.
The function $w$ is typically used to reduce the impact of patches with ambiguous content (e.g. with uniform intensities), or can be chosen to encode prior information on the target application.

The core idea of our method is to approximate the similarity metric $s(P_1, P_2)$ of two image patches with a dot product $\dotp{\phi(P_1)}{\phi(P_2)}$ where $\phi(\cdot)$ is a function that extracts a feature vector, for instance in $\mathbb{R}^{16}$, from its input patch.
When $\phi$~is a fully convolutional neural network (CNN), we can simply feed it the entire volume in order to pre-compute the feature vectors of every voxel with a single forward pass.
The registration objective (Eq.~\ref{reg_obj}) is then approximated as
\begin{align}
\label{approx}
f(\alpha) \approx \sum_{p \in \Omega} w(p) \ \dotp{\phi(F)[p]}{\phi(M) \circ T_\alpha [p]} \, ,
\end{align}
thus converting the original problem into a registration of pre-computed feature maps using a simple and differentiable dot product similarity.
This approximation is based on the assumption that the CNN is approximately equivariant to the transformation, i.e. $\phi(M \circ T_\alpha)  [p] \approx \phi(M) \circ T_\alpha [p]$.
Our experiments show that this assumption (implicitly made also by other descriptors like MIND) does not present any practical impediment.
Our method exhibits a large capture range and can converge over a wide range of rotations and deformations. 

\smallvspace

\noindent\textbf{Advantages} \ 
In contrast to many existing methods, our approach doesn't require any ground truth registration and can be trained using patches from unregistered pairs of images. This is particularly important for multi-modal deformable registration as ground truths are harder to define, especially on ultrasound.
The simplicity of our training objective allows the use of a CNN with a limited number of parameters and a small receptive field. This means that the CNN has a negligible computational cost and can generalize well across anatomies and modalities: a single network can be used for all types of images and does not need to be retrained for a new task.
Furthermore, the objective function (Eq.~\ref{approx}) can be easily differentiated without backpropagating the gradient through the CNN. This permits efficient gradient-based optimization, even when the original metric is either non-differentiable or costly to differentiate.
Finally, we quantize the feature vectors to 8-bit precision further increasing the computational speed of registration without impacting accuracy.

%% file: training.tex
\begin{figure}[t!]
    \centering
    \includegraphics[width=0.95\linewidth]{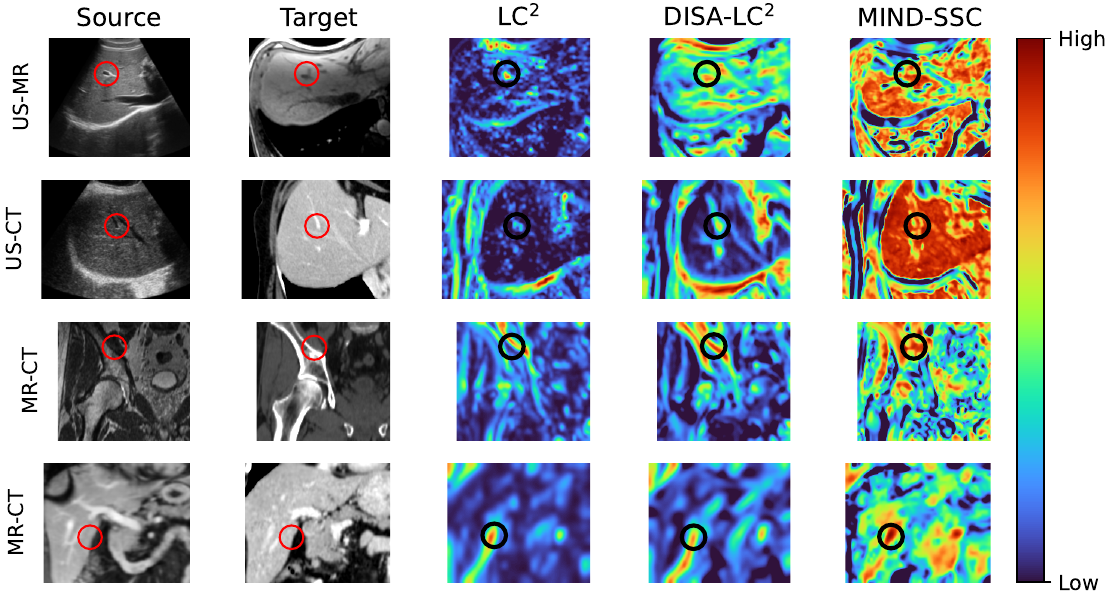}
    \caption{Similarity maps across different modalities and anatomies. Each heatmap shows the similarity of the marked point on the source image to every point in the target image.
    Our method (DISA-LC$^2$) approximates LC$^2$ well in a fraction of the computation time and produces less ambiguous heatmaps than MIND.}
    \label{fig:heatmaps}
\end{figure}
\label{sec:training}

We train our model to approximate the three-dimensional LC$^2$ similarity, as it showed good performance on a number of tasks, including ultrasound~\cite{lc2,fuerst2014automatic}.
The LC$^2$ similarity quantifies whether a target patch can be approximated by a linear combination of the intensities and the gradient magnitude of the source patch.
In order to reduce the sensitivity on the scale, our target is actually the average LC$^2$ over different radiuses of 3, 5, and 7. 
In order to be consistent with the original implementation of LC$^2$ we use the same weighting function $w$ based on local patch variance.
Note that the network will be trained only once, on a fixed dataset that is fully independent of the datasets that will be used in the evaluation (see Section~\ref{sec:experiments}).

\smallvspace

\noindent\textbf{Dataset}
Our neural network is trained using patches from the \qot{Gold Atlas - Male Pelvis - Gentle Radiotherapy}~\cite{nyholm2017_pelvis_dataset} dataset, which is comprised of 18 patients each with a CT, MR T1, and MR T2 volumes.
We resample each volume to a spacing of $2\mm$ and normalize the voxel intensities to have zero mean and standard variation of one. Since our approach is unsupervised, we don't make use of the provided registration but leave the volumes in their standard DICOM orientation. As LC$^2$ requires the usage of gradient magnitude in one of the modalities, we randomly pick it from either CT or MR. \\
We would like to report that, initially, we also made use of a proprietary dataset including US volumes.
However, as our investigation progressed, we observed that the incorporation of US data did not significantly contribute to the generalization capabilities of our model. Consequently, for the purpose of ensuring reproducibility, all evaluations presented in this paper exclusively pertain to the model trained solely on the public MR-CT dataset.

\smallvspace

\noindent\textbf{Patch sampling from unregistered datasets} 
For each pair of volumes $(M, F)$ we repeat the following procedure 5000 times:
(1) Select a patch from $M$ with probability proportional to its weight $w$;
(2) Compute the similarity with all the patches of $F$;
(3) Uniformly sample $t \in [0, 1]$;
(4) Pick the patch of $F$ with similarity score closest to $t$. Running this procedure on our training data results in a total of 510000 pairs of patches.

\smallvspace
\noindent\textbf{Architecture and Training}
We use the same feed-forward 3D CNN to process all data modalities.
The proposed model is composed of residual blocks~\cite{resnet}, LeakyReLU activations~\cite{leaky_relu} and uses BlurPool~\cite{zhang2019shiftinvar} for downsampling, resulting in a total striding factor of 4.
We do not use any normalization layer, as this resulted in a reduction in performance.
The output of the model is 16-channels volume  with the norm of each voxel descriptor clipped at 1.
The architecture consists of ten layers and a total of 90,752 parameters, making it notably smaller than many commonly utilized neural networks. 

Augmentation on the training data is used to make the model as robust as possible while leaving the target similarity unchanged.
In particular, we apply the same random rotation to both patches, randomly change the sign and apply random linear transformation on the intensity values.
We train our model for 35 epochs using the L2 loss and batch size of 256. 
The training converges to an average patch-wise L2 error of $0.0076$ on the training set and $0.0083$ on the validation set.
The total training time on an NVIDIA RTX4090 GPU is 5 hours, and inference on a $256^3$ volume takes $70\ms$.
We make the training code and preprocessed data openly available online \footnote[1]{https://github.com/ImFusionGmbH/DISA-universal-multimodal-registration}.

%% file: experiments.tex
\begin{table}[t]
\scriptsize
\centering
\renewcommand\arraystretch{1.25}
\setlength{\tabcolsep}{1.5mm}
\begin{tabular}{|lc|c|ccc|}
\hline
\textbf{Method} & \textbf{Mode} & \textbf{Avg. FRE} & \textbf{FRE25} & \textbf{FRE50} & \textbf{FRE75} \\ 
\hline
MIND-SSC & Rigid & 5.05 & 1.69 & 2.20 & 3.31\\ 
MIND-SSC & Affine & 2.01 & 1.44 & 1.84 & 2.29\\ 
\hline
LC$^2$ & Rigid & 1.71 & 1.31 & 1.56 & 1.72 \\
LC$^2$ & Affine & 1.73 & 1.32 & 1.67 & 1.89 \\
\hline
DISA-LC$^2$ & Rigid & 1.82 & 1.37 & 1.65 & 1.80 \\
DISA-LC$^2$ & Affine & 1.74 & 1.33 & 1.58 & 1.73 \\
\hline
\end{tabular}
\vspace{0.5\baselineskip}
\caption{Results on registration of brain US-MR data from the RESECT Challenge. FRE is the average of fiducial errors in millimeters across all cases, while FRE25, FRE50, and FRE75 refer to the 25th, 50th, and 75th percentiles. \label{table:brain}}
\end{table}

We present an evaluation of our approach across tasks involving diverse modalities and anatomies. 
Notably, the experimental data utilized in our analysis differs significantly from our model's training data in terms of both anatomical structures and combination of modalities.
To assess the effectiveness of our method, we compare it against LC$^2$, which is the metric we approximate, and MIND-SSC~\cite{ssc}.
In all experiments, we use a Wilcoxon signed-rank test with p-value $10^{-2}$ to establish the significance of our results.

As will be demonstrated in the next subsections, our method is capable of achieving comparable levels of accuracy as LC$^2$ while retaining the speed and flexibility of MIND-SSC.
In particular, on abdominal US registration (Section \ref{sec:ctus}) our method obtains a significantly larger capture range, opening new possibilities for tackling this challenging problem.

\subsection{Affine Registration of Brain US-MR}
In this experiment, we evaluate the performance of different methods for estimating affine registration of the REtroSpective Evaluation of Cerebral Tumors~(RESECT) MICCAI challenge dataset~\cite{resect}. This dataset consists of 22 pairs of pre-operative brain MRs and intra-operative ultrasound volumes. The initial pose of the ultrasound volumes exhibits an orientation close to the ground truth but can contain a significant translation shift.
For both MIND-SSC and DISA-LC$^2$, we resample the input volumes to $0.4\mm$ spacing and use the BFGS~\cite{BFGS_nonsmooth} optimizer with 500 random initializations within a range of $\pm 10$\textdegree and $\pm 25\mm$. \\
We report the obtained Fiducial Registration Errors (FRE) in Table \ref{table:brain}.
DISA-LC$^2$ is significantly better than MIND-SSC while the difference with LC$^2$ is not significant.
In conclusion, our experiments demonstrate that the proposed DISA-LC$^2$, combined with a simple optimization strategy, is capable of achieving equivalent performance to manually tuned LC$^2$. 

\begin{table}[t!]
\scriptsize
\centering
\renewcommand\arraystretch{1.25}
\setlength{\tabcolsep}{1.5mm}
\begin{tabular}{|lc|ccc|c|}
\hline
\textbf{Method} & \textbf{Stride} & \textbf{DSC25} & \textbf{DSC50} & \textbf{DSC75} & \textbf{HD95} \\ 
\hline
MIND-SSC & 4 & 42.3\% & 70.9\% & 84.9\% & 26.4\mm\\ 
MIND-SSC & 2 & 49.8\% & 70.9\% & 84.9\% & 24.8\mm\\ 
MIND-SSC & 1 & 48.8\% & 70.9\% & 84.9\% & 24.5\mm\\ 
\hline
DISA-LC$^2$ & 4 & 61.4\% & 72.7\% & 85.2\% & 23.6\mm\\ 
DISA-LC$^2$ & 2 & \textbf{61.5\%} & 73.2\% & 85.5\% & 22.8\mm \\ 
DISA-LC$^2$ & 1 & \textbf{61.5\%} & \textbf{74.0\%} & \textbf{85.5\%} & \textbf{22.6\mm} \\
\hline 
\end{tabular}
\vspace{0.5\baselineskip}
\caption{Results on the Abdomen MR-CT task of the Learn2Reg challenge 2021. The best results and the ones not significantly different from them are in bold.\label{table:l2r}}
\end{table}

\subsection{Deformable Registration of Abdominal MR-CT}

Our second application is the Abdomen MR-CT task of the Learn2Reg challenge 2021~\cite{learn2reg}.
The dataset comprises 8 sets of MR and CT volumes, both depicting the abdominal region of a single patient and exhibiting notable deformations.
We estimate dense deformation fields using the methodology outlined in~\cite{dense_deform} (without inverse consistency) which first estimates a discrete displacement using explicit search and then iteratively enforces global smoothness. 
Segmentation maps of anatomical structures are used to measure the quality of the registration. In particular, we compute the 25th, 50th, and 75th quantile of the Dice Similarity Coefficient (DSC) and the 95th quantile of the Hausdorff distance (HD95) between the registered label maps.  
We compare MIND-SCC and DISA-LC$^2$ used with different strides and followed by a downsampling operation that brings the spacing of the descriptors volumes to $8\mm$. 
The hyperparameters of the registration algorithm have been manually optimized for each approach.
Table~\ref{table:l2r} shows that our method obtains significantly better results than MIND-SCC on the DSC metrics while being not significantly better on HD95.

%% file: ctus.tex
\begin{figure}[t!]
    \centering
    \includegraphics[width=0.7\linewidth]{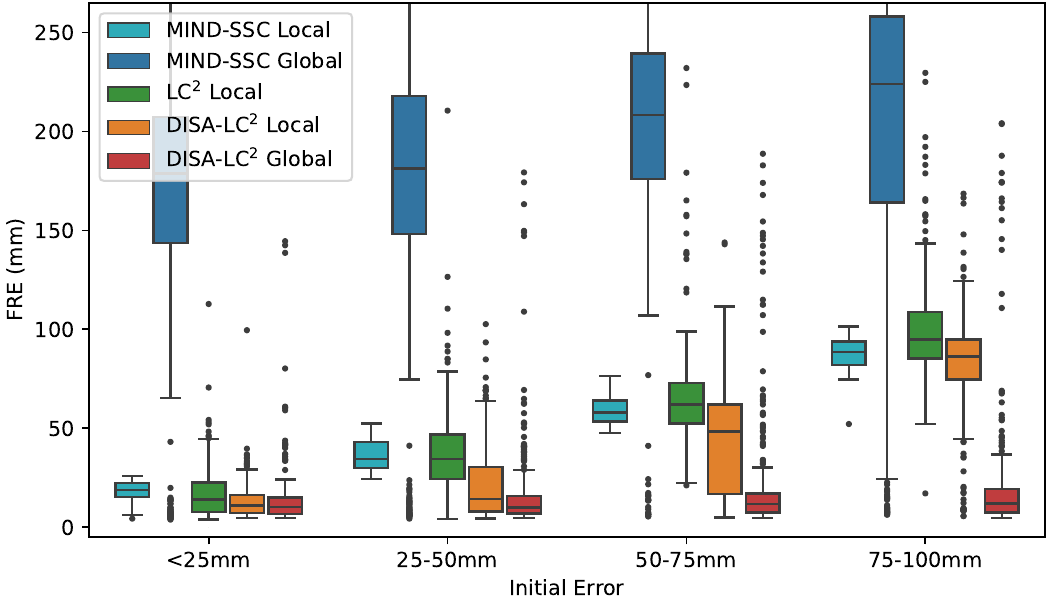}
    \vspace{-.5\baselineskip}
    \caption{Boxplot of fiducial registration errors for the different methods on deformable registration of abdominal US-CT and US-MR.}
    \label{fig:capture_range_boxplot}
\end{figure}

\begin{table}[t!]
\scriptsize
\centering
\renewcommand\arraystretch{1.25}
\setlength{\tabcolsep}{1.5mm}
\begin{tabular}{|ll|cccc|c|c|}
\hline
& & \multicolumn{4}{c|}{\textbf{Converged cases w.r.t. initialization erorr}} & \textbf{Time} & \textbf{Num.}\\ 
\textbf{Similarity} & \textbf{Search} & \textbf{0-25mm} & \textbf{25-50mm} & \textbf{50-75mm} & \textbf{75-100mm} & \textbf{(s)} & \textbf{eval.} \\ 
\hline
MIND-SSC & Local & 23.6\% & 0.0\% & 0.0\% & 0.0\% & 0.4 & 17 \\
LC$^2$ & Local & 54.1\% & 14.0\% & 0.0\% & 0.0\% & 1.9 & 98\\
DISA-LC$^2$ & Local & \textbf{70.3\%} & 52.0\% & 21.1\% & 5.8\% & 0.9 & 70 \\
\hline
MIND-SSC & Global & 17.9\% & 14.6\% & 5.3\% & 12.0\% & 1.3 & 26370\\
LC$^2$ & Global & \multicolumn{4}{c|}{N/A} & 948.0* & 38740*\\
DISA-LC$^2$ & Global & \textbf{75.5\%} & \textbf{73.2\%} & \textbf{65.0\%} & \textbf{64.0\%}  & 1.8 & 29250 \\
\hline
\end{tabular}
\vspace{0.5\baselineskip}
\caption{
Results on deformable registration of abdominal US-CT and US-MR. A case is considered \qot{converged} if the FRE after registration is less than $15\mm$. The best results and the ones not significantly different from them are highlighted in bold. (*)Time and evaluations for Global LC$^2$ are estimated by extrapolation.
\label{table:ctus}
}
\end{table}

As the most challenging experiment, we finally use our method to achieve deformable registration of abdominal 3D freehand US to a CT or MR volume. \\
We are using a heterogeneous dataset of 27 cases, comprising liver cancer patients and healthy volunteers, different ultrasound machines, as well as optical vs. electro-magnetic external tracking, and sub-costal vs. inter-costal scanning of the liver. All 3D ultrasound data sets are accurately calibrated, with overall system errors in the range of commercial ultrasound fusion options. Between 4 and 9 landmark pairs (vessel bifurcations, liver gland borders, gall bladder, kidney) were manually annotated by an expert.
In order to measure the capture range, we start the registration from 50 random rigid poses around the ground truth and calculate the Fiducial Registration Error (FRE) after optimization.
For local optimization, LC$^2$ is used in conjunction with BOBYQA~\cite{bobyqa} as in the original paper~\cite{lc2}, while MIND-SCC and DISA-LC$^2$ are instead used with BFGS.
Due to an excessive computation time, we don't do global optimization with $LC^2$ while with other methods we use BFGS with 500 random initializations within a range of $\pm 40\degree$ and $\pm 150\mm$.
We use six parameters to define the rigid pose and two parameters to describe the deformation caused by the ultrasound probe pressure.
\\
From the results shown in Table \ref{table:ctus} and Figure \ref{fig:capture_range_boxplot}, it can be noticed that the proposed method obtains a significantly larger capture range than  MIND-SCC and LC$^2$ while being more than $300$ times faster per evaluation than LC$^2$ (the times reported in the table include not just the optimization but also descriptor extraction).
The differentiability of our objective function allows our method to converge in fewer iterations than derivative-free methods like BOBYQA.
Furthermore, the evaluation speed of our objective function allows us to exhaustively search the solution space, escaping local minima and converging to the correct solution with pose and deformation parameters at once, in less than two seconds.

Note that this registration problem is much more challenging than the prior two due to difficult ultrasonic visibility in the abdomen, strong deformations, and ambiguous matches of liver vasculature. Therefore, to the best of our knowledge, these results present a significant leap towards reliable and fully automatic fusion, doing away with cumbersome manual landmark placements.

%% file: conclusions.tex
We have discovered that a complex patch-based similarity metric can be approximated with feature vectors from a CNN with particularly small architecture, using the same model for any modality. The training is unsupervised and merely requires unregistered data. After features are extracted from the volumes, the actual registration comprises a simple iterative dot-product computation, allowing for global and derivative-based optimization. This novel combination of classical image processing and machine learning elevates multi-modal registration to a new level of performance, generality, but also algorithm simplicity.

We demonstrate the efficiency of our method on three different use cases with increasing complexity. In the most challenging scenario, it is possible to perform global optimization within seconds of both pose and deformation parameters, without any organ-specific distinction or successive increase of parameter sizes.

While we specifically focused on developing an unsupervised and generic method, a sensible extension would be to specialize our method by including global information, such as segmentation maps, into the approximated measure or by making use of ground-truth registration during training.
Finally, the cross-modality feature descriptors produced by our model could be exploited by future research for tasks different from registration such as modality synthesis or segmentation.